\documentclass[conference]{IEEEtran}
\IEEEoverridecommandlockouts
\usepackage[square,sort,comma,numbers]{natbib}
\usepackage[hidelinks]{hyperref}
\usepackage{multirow}
\usepackage{amsmath,amssymb,amsfonts}
\usepackage[nameinlink]{cleveref}
\usepackage{algorithmic}
\usepackage{graphicx}
\usepackage{textcomp}
\usepackage[table]{xcolor}
\usepackage{makecell}
\usepackage{microtype}
\usepackage{xspace}

\newcolumntype{P}[1]{>{\centering\arraybackslash}p{#1}}
\newcolumntype{M}[1]{>{\centering\arraybackslash}m{#1}}
\def\BibTeX{{\rm B\kern-.05em{\sc i\kern-.025em b}\kern-.08em
    T\kern-.1667em\lower.7ex\hbox{E}\kern-.125emX}}

\newcommand{\grayrow}{\rowcolor[gray]{0.9}}

\makeatletter
\newcommand{\ie}{\emph{i.e.,}\@ifnextchar.{\!\@gobble}{}}
\newcommand{\eg}{\emph{e.g.,}\@ifnextchar.{\!\@gobble}{}}
\newcommand{\etc}{etc\@ifnextchar.{}{.\@}}
\makeatother

\DeclareMathOperator*{\argmaxA}{arg\,max} 
\DeclareMathOperator*{\argminA}{arg\,min}

\newcommand{\hide}[1]{}

\newcommand{\bdmath}{\begin{dmath}}
\newcommand{\edmath}{\end{dmath}}
\newcommand{\beq}{\begin{equation}}
\newcommand{\eeq}{\end{equation}}
\newcommand{\bdm}{\begin{displaymath}}
\newcommand{\edm}{\end{displaymath}}
\newcommand{\bea}{\begin{eqnarray}}
\newcommand{\eea}{\end{eqnarray}}
\newcommand{\beal}{\beq \begin{array}{ll}}
\newcommand{\eeal}{\end{array} \eeq}
\newcommand{\beas}{\begin{eqnarray*}}
\newcommand{\eeas}{\end{eqnarray*}}
\newcommand{\ba}{\begin{array}}
\newcommand{\ea}{\end{array}}
\newcommand{\bit}{\begin{itemize}}
\newcommand{\eit}{\end{itemize}}
\newcommand{\ben}{\begin{enumerate}}
\newcommand{\een}{\end{enumerate}}

\newcommand{\SO}{\mathrm{SO}}

\newcommand{\Real}{\mathbb{R}}

\newcommand{\SEthree}{\ensuremath{\mathrm{SE}(3)}\xspace}

\newcommand{\RPE}{\ensuremath{\mathrm{RPE}}\xspace}
\newcommand{\ATE}{\ensuremath{\mathrm{ATE}}\xspace}
\newcommand{\RMSE}{\ensuremath{\mathrm{RMSE}}\xspace}
\newcommand{\Ed}{\ensuremath{\mathrm{E}}\xspace}
\newcommand{\SOthree}{\ensuremath{\SO(3)}\xspace}

\newcommand{\T}{\mathtt{T}}
\newcommand{\R}{\mathtt{R}}

\newcommand{\residual}{\mathbf{r}}
\newcommand{\trnsp}{\mathsf{T}}

\newcommand{\rotvel}{\boldsymbol\omega}
\newcommand{\acc}{\mathbf{a}}

\newcommand{\tran}{\mathbf{p}}

\newcommand{\vel}{\mathbf{v}}

\newcommand{\bias}{\mathbf{b}}

\newcommand{\gravity}{\mathbf{g}}
\newcommand{\noise}{\boldsymbol\eta}

\newcommand{\logmap}{\mathrm{Log}}

\newcommand{\States}{\mathcal{X}}

\newcommand{\covprior}{\mathbf{\Sigma}}

\newcommand{\World}{\text{W}}
\newcommand{\Imu}{\text{B}}

\newcommand{\imu}{\text{\tiny{B}}}

\newcommand{\preintRmeas}{\Delta\tilde\R}
\newcommand{\preintVmeas}{\Delta\tilde\vel}
\newcommand{\preintPmeas}{\Delta\tilde\tran}

\begin{document}

\title{Pose Graph Optimization for a MAV Indoor Localization Fusing 5GNR TOA with an IMU
\thanks{This work was funded by the Luxembourg National Research Fund (FNR) 5G-SKY project
(ref. C19/IS/13713801) and by the European Commission Horizon2020 research and innovation
programme under the grant agreement No 101017258 (SESAME).}
}




\author{\IEEEauthorblockN{Meisam Kabiri\IEEEauthorrefmark{1},
Claudio Cimarelli\IEEEauthorrefmark{1}, Hriday Bavle\IEEEauthorrefmark{1},
Jose Luis Sanchez-Lopez\IEEEauthorrefmark{1}, and Holger Voos\IEEEauthorrefmark{1}\IEEEauthorrefmark{2} }

 \IEEEauthorblockA{\IEEEauthorrefmark{1} \textit{Interdisciplinary Center for Security Reliability and Trust (SnT)} \\
\textit{University of Luxembourg, Luxembourg}\\
\IEEEauthorrefmark{2} \textit{Faculty of Science, Technology, and Medicine (FSTM), Department of Engineering}\\
\textit{University of Luxembourg, Luxembourg}\\
Email: \{meisam.kabiri, claudio.cimarelli, hriday.bavle, joseluis.sanchezlopez, holger.voos\}@uni.lu
}}
\maketitle

\maketitle

\begin{abstract}
This paper explores the potential of 5G new radio (NR) Time-of-Arrival (TOA) data for indoor drone localization under different scenarios and conditions when fused with inertial measurement unit (IMU) data. Our approach involves performing graph-based optimization to estimate the drone's position and orientation from the multiple sensor measurements. Due to the lack of real-world data, we use Matlab 5G toolbox and QuaDRiGa (quasi-deterministic radio channel generator) channel simulator to generate TOA measurements for the EuRoC MAV indoor dataset that provides IMU readings and ground truths 6DoF poses of a flying drone. Hence, we create twelve sequences combining  three predefined indoor scenarios setups of QuaDRiGa with 2 to 5 base station antennas. Therefore, experimental results demonstrate that, for a sufficient number of base stations and a high bandwidth 5G configuration, the pose graph optimization approach achieves accurate drone localization, with an average error of less than 15 cm on the overall trajectory. Furthermore, the adopted graph-based optimization algorithm is fast and can be easily implemented for onboard real-time  pose tracking on a micro aerial vehicle (MAV).

\end{abstract}


\section{Introduction}

Drones, such as micro aerial vehicles (MAVs), have become increasingly prevalent in indoor environments due to their potential in various applications, from surveillance to delivery services. For many of these applications, accurate drone positioning and orientation are essential. 

Global Navigation Satellite Systems (GNSS), the most widely used positioning technology, encounters challenges in penetrating indoor environments due to signal attenuation and multipath effects. 
Inertial navigation systems (INSs) are another widely used method for indoor localization, but they can accumulate noise over time, resulting in significant position errors if left uncorrected.

Alternative indoor localization techniques, including Wi-Fi, Bluetooth, and Ultra-Wideband (UWB), exhibit limitations in accuracy, scalability, energy efficiency, and cost~\cite{yang2021survey, zafari2019survey}. For instance, Wi-Fi is highly susceptible to noise, Bluetooth has restricted range and accuracy, and UWB faces slow progress in standard development.
Moreover, Zigbee's emphasis on low-power communication and its limited range further constrain its localization capabilities. 
Consequently, a need for alternative indoor positioning technologies arises that can offer high accuracy and reliability without relying on GNSS.


Recent advances in wireless communication technologies have paved the way for developing location-based services and applications that rely on accurate localization. In particular, deploying fifth-generation (5G) cellular networks has opened up new possibilities for indoor localization due to their high bandwidth, low latency, and improved coverage~\cite{Kabiri2022}, with small cell technologies, like femtocells and picocells, facilitating indoor coverage.  For downlink positioning, 5G utilizes a dedicated pilot signal called Positioning Reference Signal (PRS), which measures signal delay by correlating the received PRS with a locally generated PRS. The delay, also called Time-of-Arrival (TOA), is then calculated by identifying the peak correlation value between the two signals.


Accurate and reliable indoor localization remains challenging due to the complex and dynamic nature of indoor environments. In this context, our work proposes a novel approach for indoor localization using 5G TOA measurements. However, 5G TOA alone may not provide sufficient information for reliable indoor localization. 
Thus, our work aims to fuse 5G TOA with an inertial measurement unit (IMU), to improve the real-time pose estimation of a flying MAV. IMUs can measure angular velocities, and linear acceleration, providing complementary information to TOA measurements. To accomplish this, we extract distances from the PRS correlation profiles and use them to formulate a range error function tightly integrated with the IMU measurements in a graph-based optimization technique. Due to the lack of real 5G data recorded from MAVs, we utilize the QuaDRiGa simulator~\cite{jaeckel2017quadriga} to create accurate 5G TOA measurements based on ground truth 6 Degrees of Freedom (DoF) pose data to ensure precise channel modeling. Hence, our experimental simulations include three indoor configurations with LOS and varying base stations (BSs) numbers virtually placed inside the Euroc MAV dataset environment~\cite{Burri2016}.

To summarize, the contributions provided by this paper are the following:
\begin{itemize}
    
    \item Formulation of a factor graph model that tightly optimizes 5G TOA ranges with IMU measurements. 
    \item Evaluation in a state-of-the-art dataset shows accuracy close to centimeter precision while running an efficient real-time algorithm.
    \item Simulation of 5G TOA comparing multiple antenna and communication settings in an indoor scenario to find the best configuration for precise localization.
    
     
\end{itemize}

\section{Related Works}
\label{sec:relatedworks}

The literature on localization using 5G is relatively limited, especially considering the mechanical aspects and sensor fusion framework. The existing literature often relies on simplistic methodologies and scenarios. \citet{ferre2019positioning} compared localization accuracy for different combinations of the 5G network configurations (center frequency, sub-carrier spacing, and PRS comb size) in terms of the Root Mean Square Error (RMSE). Also, their study considered a fixed target and employed multilateration based on PRS-derived TOA from multiple BSs. 
A study by \citet{PeralRosado2016} explored the impact of positioning performance when BSs are linearly placed along a straight roadside 5G network. They utilized Gauss-Newton optimization and simulated a 100 km/h vehicle on a highway. The study revealed an accuracy of less than 20-25 cm for a communication bandwidth of 50-100 MHz. Additionally, the researchers calculated the TOA by determining the first correlation peak between the PRS and the received signal. \citet{Saleh2022} proposed a time-based positioning by combining vehicle velocity information and 5G measurements. They evaluated their approach in a simulated urban canyon using Siradel's S\_5GChannel simulator and employed an EKF with a constant velocity model for sensor fusion. The study also analyzed the impact of the 5G geometrical setup on EKF positioning estimation. Another EKF-based positioning framework is proposed by \citet{menta2019performance}. The authors leveraged the 5G Angle of Arrival (AOA) extracted from the communication signal of BSs equipped with multi-array antennas. By utilizing this information, they achieved sub-meter accuracy in localization.
\citet{Sun2020} studied localization by combining AOA estimates from 5G BSs with TOA measurements from GNSS satellites. The authors utilized the Taylor series to linearize the mathematical model. As post-processing, they applied a moving averaging to the raw position estimates to minimize errors. Finally, \cite{klus2021neural} explored the fusion of beamformed RSS information with GNSS data using Neural Networks (NN), achieving meter accuracy. 


Unlike previous approaches, we address indoor localization using the factor graph to model the relation among non-homogenous sensor measurements. Furthermore, we leverage the advanced IMU pre-integration factor to propagate the MAV's 6 DoF pose between two lower frequency TOA measurements, obtaining 6DoF pose estimates at a higher frequency. 

\section{Methodology}
\label{sec:methodology}
This section describes the proposed drone localization approach in 5G networks. This involves the fusion of range measurements, which we obtain from the TOA of the signal from multiple BSs, with IMU data, \ie~angular velocity and linear acceleration measurements.  

\subsection{5G ToA Estimation}
To estimate the distance to 5G BSs, we must first obtain the ToA values along the robot's trajectory. Since we do not possess actual data on 5G communication, this step first requires generating the 5G signal transmitted by each base station, including PRS symbols.
Then, a channel simulator creates an impulse response that emulates the wireless channel characteristics based on specific network configurations and given receiver and transmitter positions. Convolving the transmitted signal with the impulse response replicates the effects of the transmission environment, generating the received signal.

At the receiver, the signal is correlated with the corresponding PRS of each base station resulting in a PRS correlation profile.
By analyzing it, we identify the TOA as one of the peak or local maxima. Notably, the real TOA is a value close to the best peak but often not matching it. 
We apply a heuristic selection of the first peak surpassing a global threshold that we find experimentally on the data.


\subsection{Pose Estimation}
\begin{figure}[!ht]
    \centering
    \includegraphics[width = \linewidth]{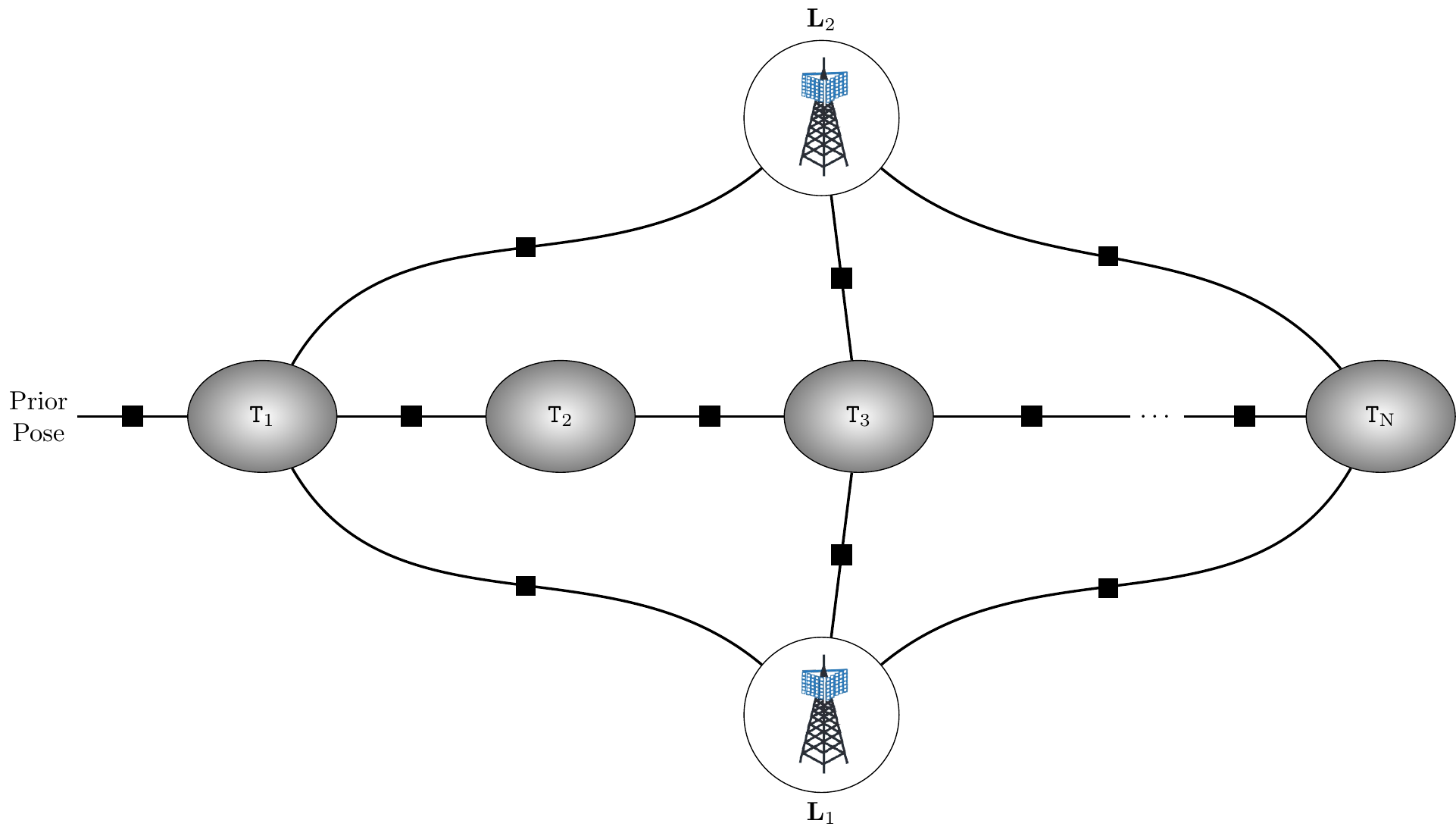}
    \caption{The figure visualizes the structure of the factor graph used to optimize the variables, represented by circles, by relating them through factors, represented by squares. The nodes $\T_t$ contain the 6DoF pose variables connected by IMU pre-integration factors (the bias and velocity nodes are not visualized). TOA measurements create range factors between robot pose nodes and BSs position nodes, $\mathbf{L}_1$ and $\mathbf{L}_2$. A prior factor is connected to the first node $\T_1$ to constrain it with the initial trajectory pose.}
    \label{fig:graph}
\end{figure}

To accurately estimate the drone's 6DoF pose, we use a graph-based optimization technique that models the relationships between the pose variables using sensor measurements and optimizes the estimation using non-linear least squares. This approach involves creating a factor graph model~\cite{factor_graphs_for_robot_perception} (see the one abstracting our problem formulation in~\autoref{fig:graph}) where the nodes represent the state variables to be estimated and the edges, called factors, represent residual error functions that compare predicted states and observed measurements. Therefore, the factor graph models the posterior probability density $p(\mathcal{X}|\mathcal{Z})$ of the state variables $\mathcal{X}$ given a set of measurements $\mathcal{Z}$, and, assuming independent measurements, it can be factorized into likelihoods $p(\mathcal{Z}^f_t|\mathcal{X}_t)$ and prior $p(\mathcal{X}_0)$:
\begin{equation}
\label{eq:fg}
    p(\mathcal{X}|\mathcal{Z}) \propto  p(\mathcal{X}_0)p(\mathcal{Z}|\mathcal{X}) = p(\mathcal{X}_0)\prod_{\forall t\in \mathcal{T}, \forall f\in \mathcal{F}} p(\mathcal{Z}^f_t|\mathcal{X}_t) \,,
\end{equation}
where $\mathcal{F}$ defines the set of factors type of functions that can replace the likelihoods. We denote the set of measurements the factor $f$ uses for computing the residual at time $t$ with $\mathcal{Z}^f_t$, where $\mathcal{T}$ is the set of tracked time frames.

\subsubsection{State Variables} 
We aim to determine the 3D location and orientation of the MAV's body center, which we align with the IMU frame. As we track the full history of 6DoF poses, with a full-smoothing approach, the set of state variables $\States$ contains all the poses from the start of the trajectory $\T_1$ to the end $\T_\mathrm{N}$, where $\mathrm{N}$ is the total number of pose nodes added to the graph. Especially, each $\T_t \doteq (\R,\tran) \in \SEthree , \forall t \in \mathcal{T}=\{ 1, \ldots , \mathrm{N}  \}$ is composed of a rotation $\R_t \in \SOthree$ and a translation $\tran_t \in \mathbb{R}^3$ that transforms the body frame $\Imu$ to the world frame $\World$, where the BSs are placed, at time $t$.

Additional to the 6DoF transformations, $\States$ comprises the MAV's linear velocity $\vel_t \in \mathbb{R}^3$. We also need to estimate the time-variant biases of the IMU's gyroscope $\mathbf{b}^g_t \in \mathbb{R}^3$ and accelerometer $\bias^a_t \in \mathbb{R}^3$ to keep track of the IMU noise drift. Lastly, we include the 3D positions of the 5G antennas $\mathbf{L}_k \in \mathbb{R}^3, \forall k \in \{1, \ldots , \mathrm{K}\}$, where $\mathrm{K}$ is the total number of BSs.

\subsubsection{IMU Factor}

Our approach involves a 6-axis IMU that measures the body $\Imu$ linear acceleration ${}_\imu\tilde\acc_t$ and angular velocity ${}_\imu\tilde\rotvel_t$ expressed in the $\World$ frame. The IMU real motion state $\{ {}_\imu\acc_t, {}_\imu\rotvel_t\}$ is altered by additive Gaussian white noise $\{\noise^{a}_t, \noise^{g}_t\}$ and slowly time-varying biases $\{\bias^a_t, \bias^g_t\}$ affecting respectively the accelerometer and gyroscope as defined by the following IMU model:
\begin{align}
  {}_\imu\tilde\rotvel_t
    &=\ {}_\imu\rotvel_t + \bias^g_t + \noise^g_t \\
  {}_\imu\tilde\acc_t
    &=\ {}_\imu\acc_t - \R_t^\trnsp \gravity + \bias^a_t + \noise^a_t \, ,
\end{align}
where $\gravity$ is the Earth's gravity vector in the world frame $\World$.

Due to the IMU's higher sampling frequency than other sensors, it typically captures multiple measurements between two TOA instances. The IMU factor is constructed utilizing a  \textit{preintegrated measurement}~\cite{imu_preintegration} constraining the relative motion increments. Especially, we obtain the condensed measurements $\preintRmeas_{ij}$ of rotation, $\preintPmeas_{ij}$ of position, and $\preintVmeas_{ij}$ of velocity by integrating multiple IMU readings $\{{}_\imu\tilde\acc_t, {}_\imu\tilde\rotvel_t : \forall t \in [i, j]\}$. So, we can define the residual terms $\residual$ for the states $\{\R_{ij}, \tran_{ij}, \vel_{ij}\}$:
\begin{align}
  \residual_{ij}^{\R}
    \doteq&\ \textstyle \logmap\left( \preintRmeas_{ij}^\trnsp\
     \R_i^\trnsp \R_j  \right)\, ,\\                  
  \residual_{ij}^{\tran}
    \doteq&\ \textstyle  \R_i^\trnsp \big( \tran_j-\tran_i - \vel_i\Delta t_{ij}
    - \frac{1}{2}\gravity \Delta t_{ij}^2\big)  -\preintPmeas_{ij}\, ,\\    
  \residual_{ij}^{\vel} 
    \doteq&\ \textstyle \R_i^\trnsp \left(\vel_j-\vel_i - \gravity\Delta t_{ij}\right)-\preintVmeas_{ij} \, ,
\end{align} 
where $\Delta t_{ij}$ is the total time interval. Also, $\logmap : \SOthree \rightarrow \Real^3$ defines the logarithm map that associates elements of the rotation manifold $\SOthree$ to vectors on the Euclidean tangent space $\Real^3$ representing rotation increments.

Regarding the biases, the total residual $\residual_{ij}^{\bias}$ between time $t=i$ and $t=j$ and $i < j$ is set as follows: 
\begin{equation}
    \residual_{ij}^{\bias} \doteq\ \bias^{g}_j - \bias^{g}_i + \bias^{a}_j - \bias^{a}_i  \, .
\end{equation}

\subsubsection{TOA Range Factor}

By multiplying the estimated TOA values $\delta_{sk}$ by the speed of light $c$, \ie~$d_{sk} = \delta_{sk} \cdot c$, 
we obtain $\mathrm{K}$ metric distance measurements $d_{sk}\in \Real, \forall s \in \mathcal{S} \subseteq \mathcal{T}$ of the drone to the $k$-th landmark $\mathbf{L}_k$ at time $s=i$. Notably, we explicitly express the possibility of having fewer TOA measurements than the number of tracked poses. The residual $\residual_{ik}^{\delta}$ of the TOA factor at time $s=i$ with the BS $\mathbf{L}_k$ is defined as: 
\begin{equation}
  \residual_{ik}^{\delta} \doteq  d_{ik} - \| \tran_i - \mathbf{L}_k\|_2 \, .
\end{equation}

\subsubsection{Optimization}

The optimization problem is formulated as Maximum a Posteriori Estimation (MAP) estimation that involves finding the state $\mathcal{X}^*$ that maximizes the posterior:
\begin{equation}
\label{eq:map0}
    \mathcal{X}^* = \argmaxA_{\mathcal{X}}  p(\mathcal{X}|\mathcal{Z}) .
\end{equation}
Considering the proportional relationship in \autoref{eq:fg}, \autoref{eq:map0} is equivalent to the maximization of the problem factorized through likelihood functions:
\begin{equation}
\label{eq:map}
    \mathcal{X}^* = \argmaxA_{\mathcal{X}} p(\mathcal{X}_0) \prod_{\forall t\in \mathcal{T}, \forall f\in \mathcal{F}} p(\mathcal{Z}^f_t|\mathcal{X}_t)\, .
\end{equation}
Notably, with factor graphs, likelihoods can be expressed by the more general factors, which we have defined with $f\in \mathcal{F} = \{ \R, \tran, \vel, \bias, \delta \}$ referring to the related residual functions.
By assuming that the measurements' errors are zero-mean Gaussian distributed, \autoref{eq:map} is analogous to the minimization of the negative log-likelihood: 
\begin{equation}
\mathcal{X}^* =\ \argminA_{\mathcal{X}} \left\|\residual_0\right\|^2_{ \mathbf{\covprior}_0} + \sum^\mathrm{N}_{t=1} \sum^{}_{\forall f\in{\mathcal{F}}}
\left\|\residual_t^f\right\|^2_{ \mathbf{\covprior}_t^f} \, ,
\end{equation}

where $\left \| \mathbf{r} \right\|_\mathbf{\covprior}^2 = \mathbf{r}^\trnsp \mathbf{\covprior}^{-1}{\mathbf{r}}$ is the squared Mahalanobis norm, and $\residual_t^f$ are the residual functions of the aforementioned factors $f$ computed at time $t$ with covariance matrix $\mathbf{\covprior}_t^f$. We denote with $\residual_0$ the residual derived from the prior on the initial pose with $\mathbf{\covprior}_0$ being its covariance matrix.

To efficiently solve the MAP optimization problem, we utilize the iSAM2 iterative optimization algorithm~\cite{ISAM2} implemented in GTSAM~\cite{gtsam}. This algorithm can automatically identify the variables that require linearization at each step, and it enables us to keep our graph solution updated while adding new nodes without experiencing memory overload.


\section{Experiments}
\label{sec:experiment}

\subsection{Simulation of the 5G Communication}
This study uses 5G specifications for indoor base stations to generate PRS signals for TOA signals. Our simulation relies on the MATLAB 5G Toolbox to generate the resource grids for 5G NR signals, including the PRS and Physical Downlink Shared Channel (PDSCH) resources. 

Aiming at a realistic simulation of the 5G NR wireless communication, we employ QuaDRiGa (quasi-deterministic radio channel generator) channel simulator~\cite{jaeckel2017quadriga}. For precise channel modeling, it is essential to have information regarding the drone's position, orientation, and velocity. This information is required to account for factors such as the Doppler shift effect accurately. We also assume that both the receiver and all transmitters use omnidirectional antennas. Three 5G simulation scenarios were considered, with different frequencies: QuaDRiGa-Industrial-LOS for 5 GHz, 3GPP-38.901-Indoor-LOS for 28 GHz, and mmMAGIC-Indoor-LOS for 78 GHz.

\begin{itemize}
    \item {\bf QuaDRiGa\_Industrial\_LOS \cite{Jaeckel2019}}: This scenario is designed to replicate a LOS environment for industrial applications.

    \item {\bf 3GPP\_38.901\_Indoor\_LOS \cite{3GPP2018}}: This scenario simulates indoor environments, \eg~office buildings and shopping centers, with 0.5-100 GHz LOS frequency. 

    \item {\bf mmMAGIC\_Indoor\_LOS \cite{CarneirodeSouza2022}}: This is designed specifically for frequencies ranging from 6-100 GHz and indoor environments, \eg~offices, with LOS.


\end{itemize}

The configurations for each 5G simulation scenario are provided in~\autoref{tab:configs}.
\begin{table}[h!]
    \centering
    \caption{5G simulation scenarios' system configurations.}
    \label{tab:configs}
    \resizebox{.8\linewidth}{!}{
    \begin{tabular}{ M{0.17 \textwidth} M{0.06 \textwidth} M{0.07 \textwidth}  M{0.08 \textwidth}  M{0.05 \textwidth}  M{0.06\textwidth} M{0.06\textwidth} M{0.07\textwidth}   } 
        \Xhline{3\arrayrulewidth}
    \bf Scenario & \bf  Freq. (GHz) & \bf Bandwidth (MHz)  & \bf Subcarrier Spacing (KHz) & \bf Num. of RBs & \bf Comb Size & \bf SNR  & \bf Cyclic Prefix\\
    \Xhline{3\arrayrulewidth}
    \grayrow
    QuaDRiGa-Industrial-LOS & 5 (FR1) & 100  & 30 & 275  & 6 & 10 dB & normal \\
    
     3GPP-38.901-Indoor-LOS & 28 (FR2)& 200  & 60 & 275  & 6 & 10 dB & normal \\
    \grayrow
    mmMAGIC-Indoor-LOS & 78 (FR2)& 400  & 120  & 275 & 6 & 10 dB & normal \\
    \Xhline{3\arrayrulewidth}

    \end{tabular}
    }
\end{table}

\subsection{Experimental Environment}
As the 5G channel simulation requires knowing the state of the receiver, \ie~its pose and velocity, we require a flying drone dataset to evaluate our method. The EuRoC MAV dataset~\cite{Burri2016} is a widely used benchmark dataset for visual-inertial odometry and SLAM. It was collected by an indoor drone equipped with a stereo-camera module providing images at 20 Hz and a calibrated IMU at 200HZ. The EuRoC MAV dataset contained the drone's position and orientation data obtained through the Vicon motion capture system, which can record the full 6DoF at about 100Hz. The full set of calibrated rigid transformations between sensors and the Vicon is also given. EuRoC MAV consists of several sequences. For this study, we consider the Vicon Room 1, sequence 01.


We employ QuaDRiGa to model the wireless communication channel based on the available 6DoF ground truth poses of the drone provided by the dataset from which we compute the required velocity considering the translation vectors between two time-consecutive poses. To this aim, we virtually place two to five fictitious BSs in the room where the trajectory is recorded. The positions of the BSs in the EuRoC MAV Vicon system's coordinate frame are $\mathrm{BS}_1 = (-10, -7, 2)$, $\mathrm{BS}_2 = (7, 13, 3)$, $\mathrm{BS}_3 = (25, -35, 4)$, $\mathrm{BS}_4 = (-6, 9, 5)$, $\mathrm{BS}_5 = (-4, -14, 6)$.
We use these values to initialize the corresponding state variables of the optimization problem with a small covariance.


After generating the resource grids and simulating the channel model, we generated the received signal at the receiver every 0.2 seconds enabling the calculation of TOA with the frequency of 5 Hz. To extract the TOA, the received signal correlated with the transmitter's PRS pattern, and the delay was calculated by analyzing the correlation profile. Typically, the initial or highest peak is considered as the response. This approach can be compromised by noise. The LOS coefficient may be weaker than the multipath coefficient due to attenuation from non-line-of-sight (NLOS) objects or constructive interference. To address this, a threshold was set to eliminate values below it, and the first peak above the threshold was chosen as the response. It is worth noting that the threshold value was determined through experimentation.

\begin{figure}[!h]
    \centering
    \includegraphics[width=.95\linewidth]{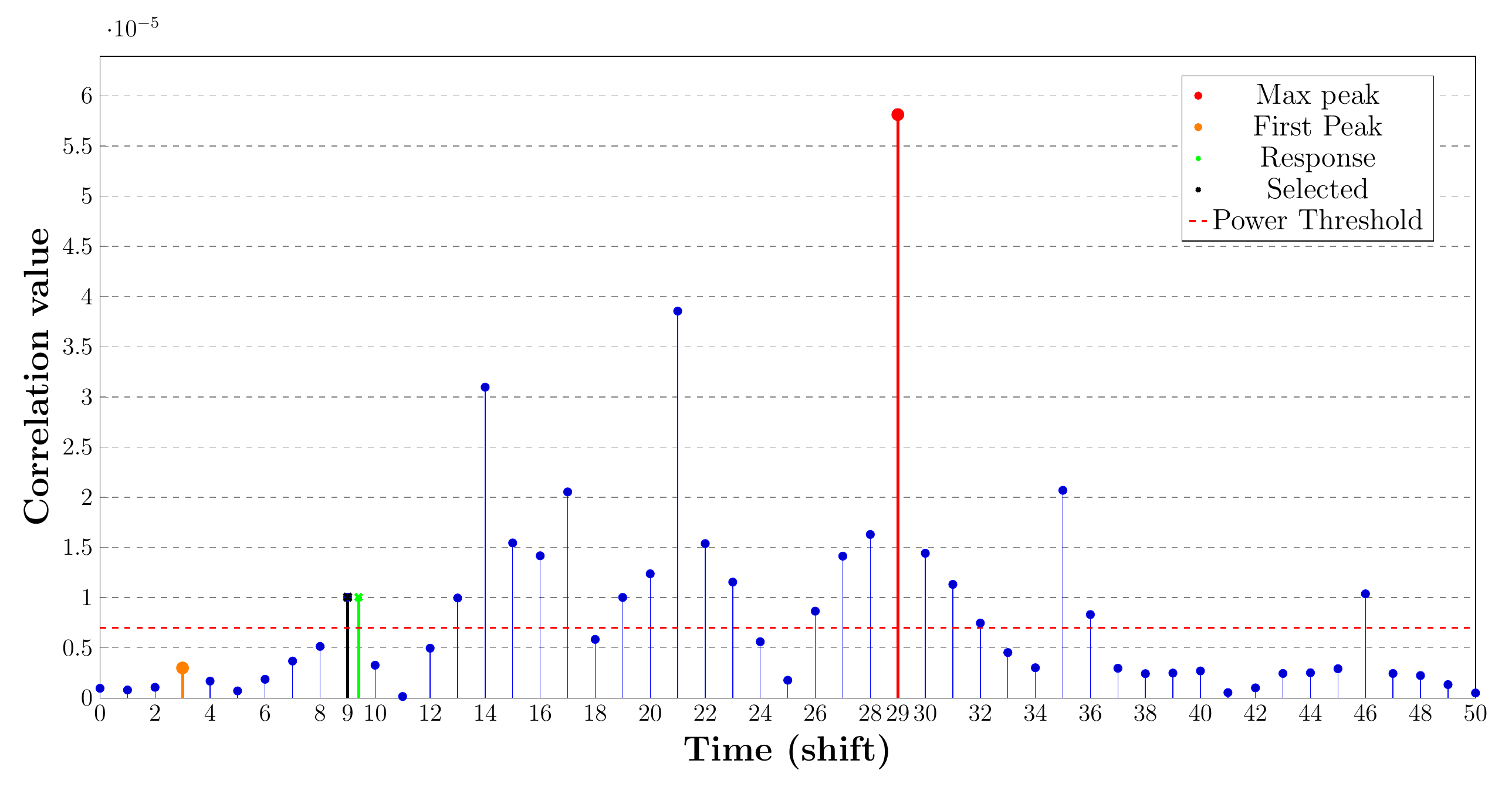}
    \caption{PRS Correlation Profile}
    \label{fig:corr_profile}
\end{figure}

An example of a correlation profile in the simulation is shown in~\autoref{fig:corr_profile}, where neither the first nor the maximum peak was the response. Still, a suitable threshold allowed the selection of the first peak as the response. 
\autoref{tab:stats} gives the statistic of the error in the resulting estimated distance to each BS.

\begin{table}[!h]
    \centering

    \caption{Error statistics of the estimated TOA distance to BSs for different 5G simulation scenarios given in meters.}
    \label{tab:stats}
    \setlength{\extrarowheight}{0.5em} 
    \resizebox{.8\linewidth}{!}{
    \begin{tabular}{c c c c c c c}
       
    \Xhline{3\arrayrulewidth}
    \bf{5G Sim. Scenario} & \bf Statistic & \bf TOA\#1 & \bf TOA\#2 & \bf TOA\#3 & \bf TOA\#4 &  \bf TOA\#5 \\
    \Xhline{3\arrayrulewidth}
    \multirow{2}{*}{ QuaDRiGa-Industrial-LOS } & \cellcolor[gray]{0.9} Mean & \cellcolor[gray]{0.9} {0.128} & \cellcolor[gray]{0.9} {-0.044} & \cellcolor[gray]{0.9} {0.005} & \cellcolor[gray]{0.9} {-0.080} & \cellcolor[gray]{0.9} {-0.022} \\
    
     & {\centering Std.} & {0.568} & {0.810} & {0.763} & {0.872} & {0.717} \\
     \Xhline{3\arrayrulewidth}
    
     \multirow{2}{*}{\centering 3GPP-38.901-Indoor-LOS} & \cellcolor[gray]{0.9}  {Mean} & \cellcolor[gray]{0.9} {-0.024} & \cellcolor[gray]{0.9} {-0.021} & \cellcolor[gray]{0.9} {-0.059} & \cellcolor[gray]{0.9} {0.040} & \cellcolor[gray]{0.9} {-0.059} \\
      & {\centering Std.} & {0.344} & {0.368} & {0.351} & {0.394} & {0.3690}   \\
      \Xhline{3\arrayrulewidth}
   
      \multirow{2}{*}{\centering mmMAGIC-Indoor-LOS } & \cellcolor[gray]{0.9} {Mean} & \cellcolor[gray]{0.9} {0. 0.0021} & \cellcolor[gray]{0.9} {0.0103} & \cellcolor[gray]{0.9} {-0.007} & \cellcolor[gray]{0.9} {0.0032}& \cellcolor[gray]{0.9} {-0.0104} \\
       & {\centering Std.} & {0.1845} & {0.1709} & {0.1728} & {0.1592}& {0.1763} \\
       \Xhline{3\arrayrulewidth}

    \end{tabular}
    }
    \end{table}
\subsection{Evaluation Metrics}

For the evaluation of our approach, we utilize the two most popular metrics in SLAM: Absolute Trajectory Error ($\ATE$) and Relative Pose Error ($\RPE$) of the rotation $\RPE_\R$ and translation $\RPE_\tran$~\cite{prokhorov2019measuring}.
In addition to these metrics, we calculated the $\RMSE$ error $\Ed_{a}, \forall a \in \left[ x, y, z\right]$ for each trajectory coordinate axis. By calculating the error for each coordinate axis separately, we aim to gain insights into possible differences in accuracy that depend on the spatial direction. 


 

\subsection{Results}

The drone's position and orientation results are obtained from the factor graph based on the final MAP estimate for each node. Nodes are generated consistently at 10Hz, twice the TOA's frequency. To evaluate the performance of the localization algorithm, error metrics are computed by comparing the ground truth Vicon pose with the estimated pose that is temporally closest. 

Based on the results, it is evident that increasing the number of BSs enhances the precision of position estimation in all situations except the first simulation scenario when transitioning from three to four. In particular, the mmMAGIC-Indoor-LOS with the highest bandwidth (400 MHz) outperforms the other 5G simulation scenarios.

\begin{figure}[!ht]
    \centering
    \includegraphics[width = .95\linewidth]{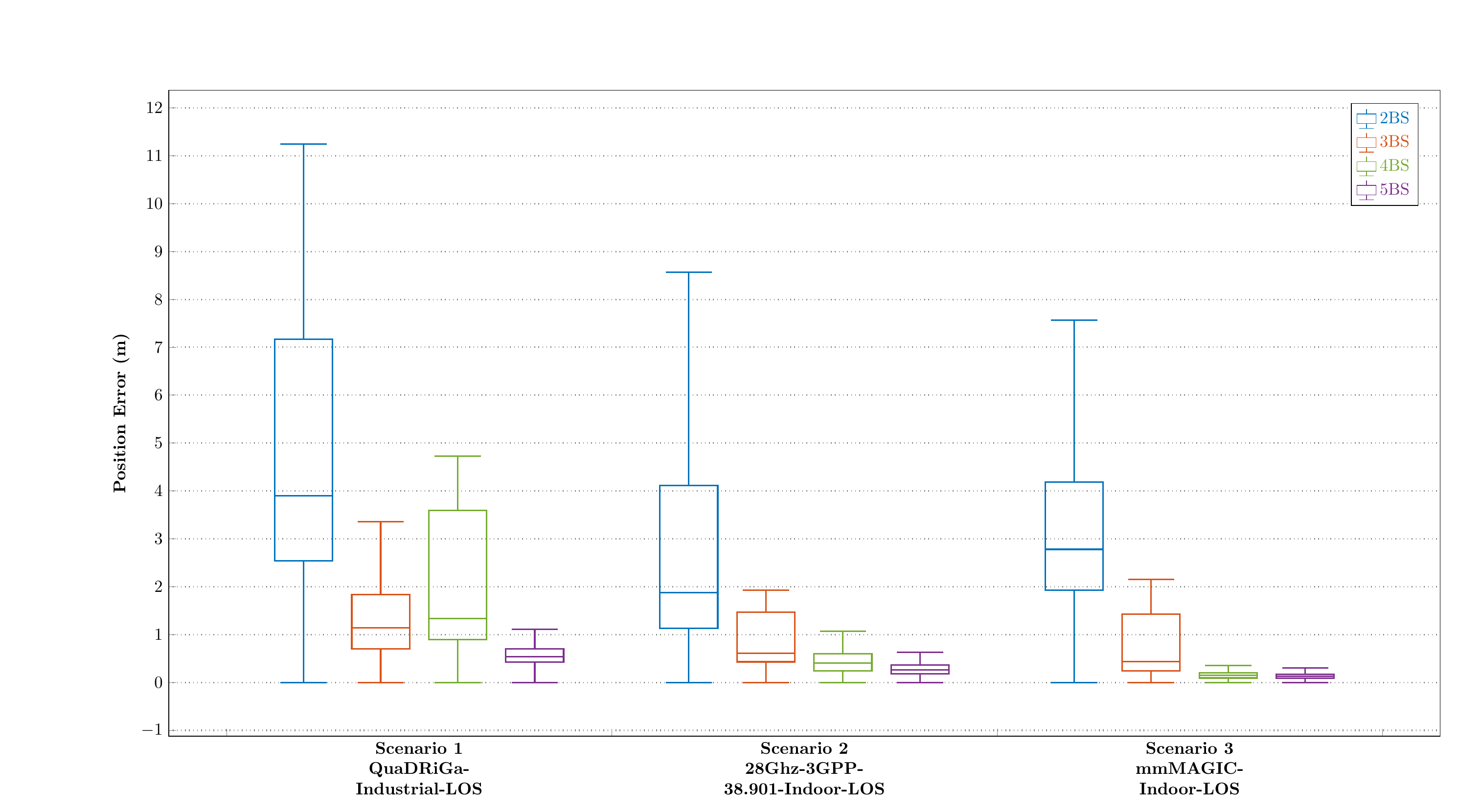}
    \caption{Box plot of the translation error in meters for each 5G simulation scenario and number of BSs.}
    \label{fig:boxplot}
\end{figure}

\autoref{fig:boxplot} illustrates the 3D position error distribution. As expected, the median value of the position error decreases with an increase in the number of BSs. The box plot reveals that the mmMAGIC-Indoor-LOS 5G simulation scenario achieves the lowest overall error. 
Surprisingly, the first scenario performs slightly worse when transitioning from three to four BSs.

\begin{table}
    \centering
    \caption{Evaluation of the pose estimation under different 5G simulation scenarios.}
    \label{tab:results}
    \resizebox{\linewidth}{!}{
    \begin{tabular}{ M{0.18 \textwidth}  M{0.04 \textwidth}  M{0.08 \textwidth}  M{0.08 \textwidth}  M{0.08 \textwidth}  M{0.08 \textwidth}  M{0.08\textwidth} M{0.08 \textwidth}  M{0.09 \textwidth}  M{0.09 \textwidth}  } 
        \Xhline{3\arrayrulewidth}
    \bf 5G sim. scenario  & \bf BS Num. & \bf $\ATE$(m) &\bf  $\Ed_x$(m) & \bf $\Ed_y$(m)& \bf $\Ed_z$(m)  & \bf $\RPE_\tran$(m) & \bf $\RPE_\R$(deg) \\
 
    \Xhline{3\arrayrulewidth}

 \multirow{4}{*}{ QuaDRiGa-Industrial-LOS } & \cellcolor[gray]{0.9} 2 & \cellcolor[gray]{0.9} 3.7290 & \cellcolor[gray]{0.9} 0.7326 & \cellcolor[gray]{0.9} 1.1775 & \cellcolor[gray]{0.9} 3.4615 & \cellcolor[gray]{0.9} 0.0174 & \cellcolor[gray]{0.9} 0.3979 \\
 & 3 & 1.2566 & 0.2978 & 0.1910 & 1.2058 & 0.0113 & 0.3984  \\
 & \cellcolor[gray]{0.9} 4 & \cellcolor[gray]{0.9} 1.5274 & \cellcolor[gray]{0.9} 0.2230 & \cellcolor[gray]{0.9} 0.1588 & \cellcolor[gray]{0.9} 1.5026 & \cellcolor[gray]{0.9} 0.0091 & \cellcolor[gray]{0.9} 0.4175  \\
 & 5 & 0.6791 & 0.2431 & 0.1553 & 0.6147 & 0.0082 & 0.4142  \\
\Xhline{3\arrayrulewidth}

 \multirow{4}{*}{ 3GPP-38.901-Indoor-LOS } & \cellcolor[gray]{0.9} 2 & \cellcolor[gray]{0.9} 2.5805 & \cellcolor[gray]{0.9} 1.0242 & \cellcolor[gray]{0.9} 1.6213 & \cellcolor[gray]{0.9} 1.7266 & \cellcolor[gray]{0.9} 0.0182 & \cellcolor[gray]{0.9} 0.3979  \\
 & 3 & 1.1189 & 0.1935 & 0.1225 & 1.0952 & 0.0083 & 0.3979 \\

 & \cellcolor[gray]{0.9} 4 & \cellcolor[gray]{0.9} 0.3787 & \cellcolor[gray]{0.9} 0.1757 & \cellcolor[gray]{0.9} 0.1147 & \cellcolor[gray]{0.9} 0.3152 & \cellcolor[gray]{0.9} 0.0070 & \cellcolor[gray]{0.9} 0.4114 \\
 & 5 & 0.2583 & 0.1717 & 0.0970 & 0.1668 & 0.0066 & 0.4487 \\
\Xhline{3\arrayrulewidth}

 \multirow{4}{*}{ mmMAGIC-Indoor-LOS } & \cellcolor[gray]{0.9} 2 & \cellcolor[gray]{0.9} 1.9861 & \cellcolor[gray]{0.9} 0.7814 & \cellcolor[gray]{0.9} 1.0087 & \cellcolor[gray]{0.9} 1.5220 & \cellcolor[gray]{0.9} 0.0104 & \cellcolor[gray]{0.9} 0.3978 \\
 & 3 & 1.2447 & 0.1228 & 0.0522 & 1.2375 & 0.0065 & 0.4378 \\

 & \cellcolor[gray]{0.9} 4 & \cellcolor[gray]{0.9} 0.1432 & \cellcolor[gray]{0.9} 0.0576 & \cellcolor[gray]{0.9} 0.0468 & \cellcolor[gray]{0.9} 0.1225 & \cellcolor[gray]{0.9} 0.0055 & \cellcolor[gray]{0.9} 0.4468 \\
 & 5 & 0.1312 & 0.0516 & 0.0422 & 0.1131 & 0.0052 & 0.4571 \\
\Xhline{3\arrayrulewidth}

    \end{tabular}
    }
\end{table}

The detailed results are given in~\autoref{tab:results}. The table includes information on ATE, RPE, and translation RMSE for each motion direction in different 5G simulation scenarios. The error values confirm that a higher number of antennas and a larger bandwidth decrease sensibly the error. However, in the last two best simulation scenarios, the accuracy improvement is marginal by increasing from four to five BSs. This may indicate a lower bound to the error reduction achieved by adding more antennas. Nevertheless, such redundancy may be helpful in those environments where NLOS conditions are more frequent.

\begin{figure}[!htpb]
    \centering
    \includegraphics[width = \linewidth]{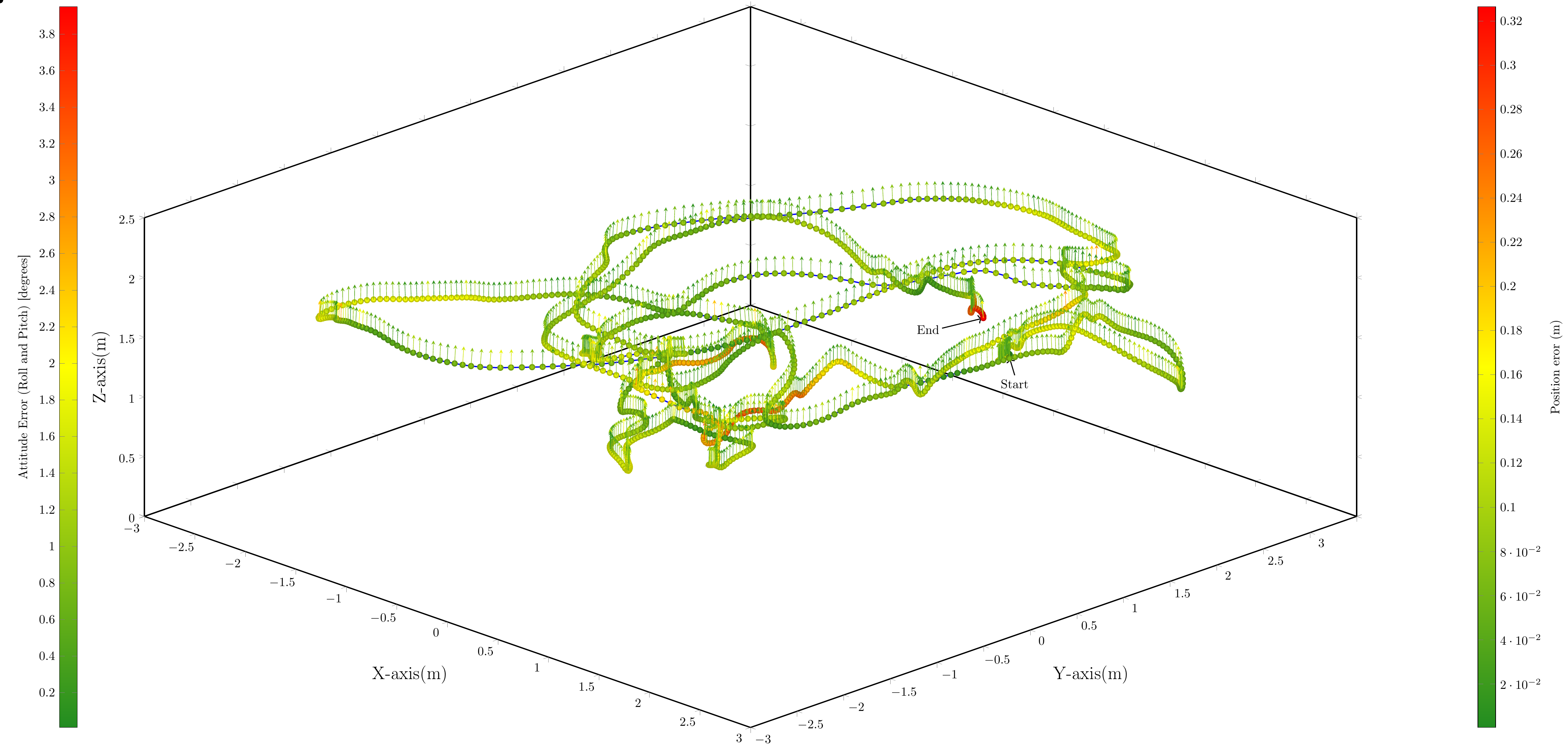}
    \caption{Visualization of the 3D trajectory estimated using five BSs in mmMAGIC-Indoor-LOS 5G simulation scenario. Best viewed online and in color.
    }
    \label{fig:3d}
\end{figure}

In~\autoref{fig:3d}, we represent the most accurate results obtained in the 5G simulation scenario mmMAGIC-Indoor-LOS with five BSs. The plot displays the estimated 3D positions of the MAV, with arrows indicating the estimated attitude (excluding yaw). The position and orientation errors are color-coded to show in red the few spots in which the poses do not match well the ground-truth and in green where the error is low, down to a few centimeters.

To evaluate the efficiency of the proposed method and its potential for real-time application, we recorded and reported the sum, average, and median optimization times. The sum optimization time was calculated to be 5.203 seconds for the 144-second trajectory of the EuRoC Mav dataset, indicating the total time required to optimize the drone's position and orientation using our graph-based framework. On average, the optimization process took only 0.0036 seconds, demonstrating the method's speed and potential for efficient implementation. Furthermore, the median optimization time was 0.0029 seconds, indicating that most optimization processes were even faster than the average, highlighting the consistency of the algorithm's performance. 

All the experiments were performed on a Ubuntu 20.04 laptop with an Intel(R) Core(TM) i9-10885H CPU @ 2.40GHz with 16 cores and 32 Gb of RAM. As the code was partially implemented in Python, we expect further improvement by a complete conversion in C++.

\subsection{Limitations}
The approach used in the study has several limitations and potential for future work. The 3D position is not fully constrained with only two antennas, making convergence difficult without other measurements. Nevertheless, the UAV's rotation errors primarily result from IMU noise, as the radio frequency signal only provides distances to the antennas. 
The yaw estimation has drift issues because it lacks global measurement to correct it. Integrating other sensors can improve the localization accuracy by observing the rotation around $z$, \eg~employing a magnetometer. Notably, a camera can be incorporated to add other constraints on the 6DoF relative motion based on visual features and loop closures.
%

Furthermore, the error in the $z$ axis is larger than along $x$ and $y$ axes because of limited offset or variation in the positions of the base stations in the height direction. We foresee the possibility of fusing the barometer's absolute height measurements to relieve such issues. Additionally, the localization accuracy depends heavily on the quality of the TOA measurements, which can be negatively affected by NLOS conditions. In such cases, correctly setting the measurement uncertainty for each TOA range factor, using Mahalanobis distance to discard outliers, or applying a robust kernel to the cost function, \eg~Huber, may be beneficial to alleviate the problem.

Finally, the proposed method assumes that the positions of the base stations are known with high confidence and fixed in the exact location, which may not be the case in real-world scenarios where the stations may be moving or their positions may be completely unknown. Moreover, the study assumes that the odometry frame can be initially aligned with the world frame inside which the antennas are placed. This can be solved in future work by explicitly estimating the transformation between the local coordinate frame and the world. 

\section{Conclusion}
\label{sec:conclusion}
In conclusion, we have successfully demonstrated the potential of using 5G TOA-based range measurements with data from inertial sensors to locate a MAV indoors in various scenarios and network setups. Our optimization strategy, which is graph-based, enables us to accurately determine the drone's position and orientation, with an average testing error of less than 15 cm. This technique has many practical applications, such as drone-powered monitoring and communication systems. In the future, we plan to improve localization accuracy and reliability by integrating visual data from cameras, experimenting with real data, and investigating advanced techniques for precise TOA estimation.


\bibliographystyle{IEEEtranN}
\bibliography{Refs}

\end{document}